# Detection of Face using Viola Jones and Recognition Using Back Propagation Neural Network

Smriti Tikoo, Nitin Malik

Department of Electrical and Electronics and Communication Engineering

The NorthCap University, Gurgaon

smrititikoo@gmail.com, nitinmalik@ncuindia.edu

## ABSTRACT

Detection and recognition of the facial images of people is an intricate problem which has garnered much attention during recent years due to its ever increasing applications in numerous fields. It continues to pose a challenge in finding a robust solution to it. Its scope extends to catering the security, commercial and law enforcement applications. Research for moreover a decade on this subject has brought about remarkable development with the modus operandi like human computer interaction, biometric analysis and content based coding of images, videos and surveillance. A trivial task for brain but cumbersome to be imitated artificially. The commonalities in faces does pose a problem on various grounds but features such as skin color, gender differentiate a person from the other. In this paper the facial detection has been carried out using Viola Jones algorithm and recognition of face has been done using Back Propagation Neural Network(BPNN).

## I. INTRODUCTION

Face of a person is unique in a manner that it possesses a set of features which might resemble with the other face in some or the other way. It is the skin color luminescence which makes a difference in the recognition of a face from a gallery of images (including the background) . Face detection is the foremost and necessary stride towards facial recognition it encompasses a variety of intermediate steps such as facial analysis algorithms, alignment, remodeling and etc. The hurdles encountered in this path can be accredited to intensity of skin color, location, size of the facial image, occlusions, pose (out of plane rotation), orientation (in plane rotation).Innumerable advances were made to cease the purpose of detecting and recognizing a facial image by the research scholars across the globe.

Schneiderman and Kanade [1] used the feature based method utilizing the geometric facial features with belief networks for frontal, profile and nonfrontal orientation of the images. L.Liao[2] incorporated Fast Fourier transform algorithms for the input face image before applying recognition. The input images of face where Fast Fourier Transform transformed to whitened faces and facial recognition was carried out using Principal Component analysis and Independent component analysis. The procedure of whitening the faces facilitated better results via Principal Component Analysis and Independent Component Analysis .Lin [3] brought about the Hausdroff distance for the recognition of faces; as a matter of fact the results were satisfactory for recognition and the processing. Sven Loncaric and Zdrvako Liposcak [4] proposed a method of recognition for profile images based on the authentic and morphological derived profile shapes, the source of information was taken from the line bounding the face and hair. Simon Ceolin et. al[5] proposed the ideas from the statiscal shape analysis for constructing shape spaces encompassing the facial features and gender and the resultant model for recognition. Ki Chung Chung [6] used the PCA and Gabor filter responses to recognize a face. We propose to detect and recognize facial images using the Back propagation neural network(BPNN) .Viola Jones algorithm has been used to detect a face and the features and recognition of face via BPNN .

## II. PROBLEM DEFINITION

The advent of biometrics brought about a change in the way security of data or people and country was treated. Be it via surveillance systems or the full body scans or the retina scans, hand scans , finger prints not only it has changed the security setup of the country but it offers a sense of reliability to the users and system as well. It has invented ways to protect and preserve the confidential data and guard the system(of any type) with help of human and computer interaction. It has brought about a great combination of image processing and computer vision to light , that in a way have boosted the business for detection and recognition systems. In this paper we propose a similar approach to detect and recognize a facial image using a BPNN with help of MATLAB 8.2. In the MATLAB we have worked using the neural network tool box, within which we have made use of the neural network fitting tool to train and test the facial image at hand. This tool maps the numeric inputs to their targets. It helps in training , testing and validating the data and evaluates the data on mean square error generated , performance plot and if needed regression plot for the same data can also be put to use to make an inference regarding the performance of the process .

## III. BACKPROPAGATION NEURAL NETWORK

It is a network which is consists of different interconnected layers. The term Back Propagation refers to the backward propagation of errors in conjunction with optimization method i.e steepest descent. It calculates the local minima with respect to the associated weights of the network. The weights are updated accordingly in order to reduce the local minima. Since this network relies on a known target output for every input fed into the network, it is thereby a supervised learning method. This algorithm is best understood by categorizing under two main phases namely:

Phase 1: Propagation:

1. Forward propagation: Input is fed through the network to generate propagation's output activations.
2. Backward propagation: A feedback network is formed by feeding the output as input in order to generate a difference between actual and the target outputs.

Phase 2: Weight update:

1. Gradient of weight is a product of difference of outputs and input activation.
2. Subtract a ratio (percentage) of the gradient from the weight.

$$f(x) = \frac{1}{1+e^{-x}}$$

Expression for the sigmoidal function used where e is the natural logarithmic function and x can have any real value .

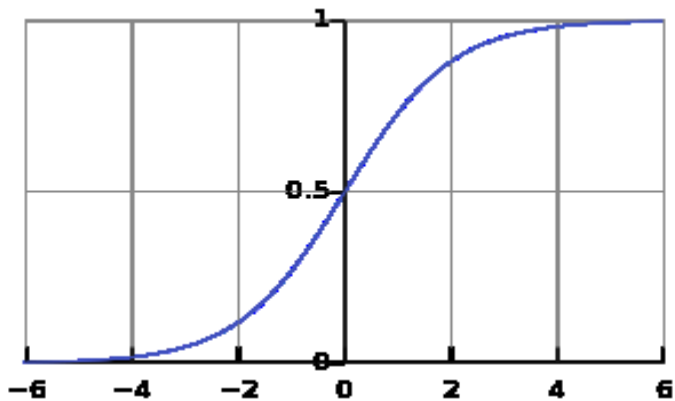

Figure sigmoidal function

It is defined by $S(t) = 1/(1 + e^{-t})$. It is a limited differentiable real function which is defined for real values and gives a positive derivative.. $S'(t) = S(t)(1 - S(t))$

## VIOLA JONES ALGORITHM

Is the first framework for object detection which gave viable results for real time situations. Paul Viola and Michael Jones had proposed the algorithm in year 2001. It was aimed at targeting the problem of face detection but can also be trained for detecting different object classes. It is implemented in Open CV as cvHaarDetectObjects ().It is preferred for its robust nature and its fast detection of faces (full frontal upright faces)in practical situations. It comprises for four stages namely:

1. Haar Feature Selection.
2. Creating an integral image.
3. Adaboost Training.
4. Cascading Amplifiers.

   Haar Feature selection matches the commonalities found in human faces. The integral image calculates the rectangular features in fixed time which benefits it over other sophisticated features.. Integral image at (x,y) coordinates gives the pixel sum of the coordinates above and on to the left of the (x,y).Ada boost training algo is used to train the classifiers and to construct a strong classifiers by cascading the previously used weak classifiers.

## IV.PROPOSED METHODOLOGY

1. Detection of face and its features using Viola Jones algorithm.
2. Conversion from rgb to grayscale and binary.
3. Segmentation of image can be done in any format.
4. Histogram equivalent of binary or grayscale image.
5. Each segmented part is trained individually using the back propagation neural network tool.
6. The above mentioned steps can also be carried out without segmentation of the image.
7. For recognition process neural network tool in Matlab is used to train and test the sample data provided by the histogram equivalent of the image taken.
8. In neural network tool, curve fitting tool has been made use to train the data.
9. The training of data takes place in 3 stages mainly training, testing and validation.
10. The mean squared error is a parameter which is focused to be able to contemplate whether the training is giving desirable results or not.
11. The data is retrained to achieve better performance plots with least mean squared error.
12. The training phase continues with usage of different samples in order to check the performance of the network and to get better results.

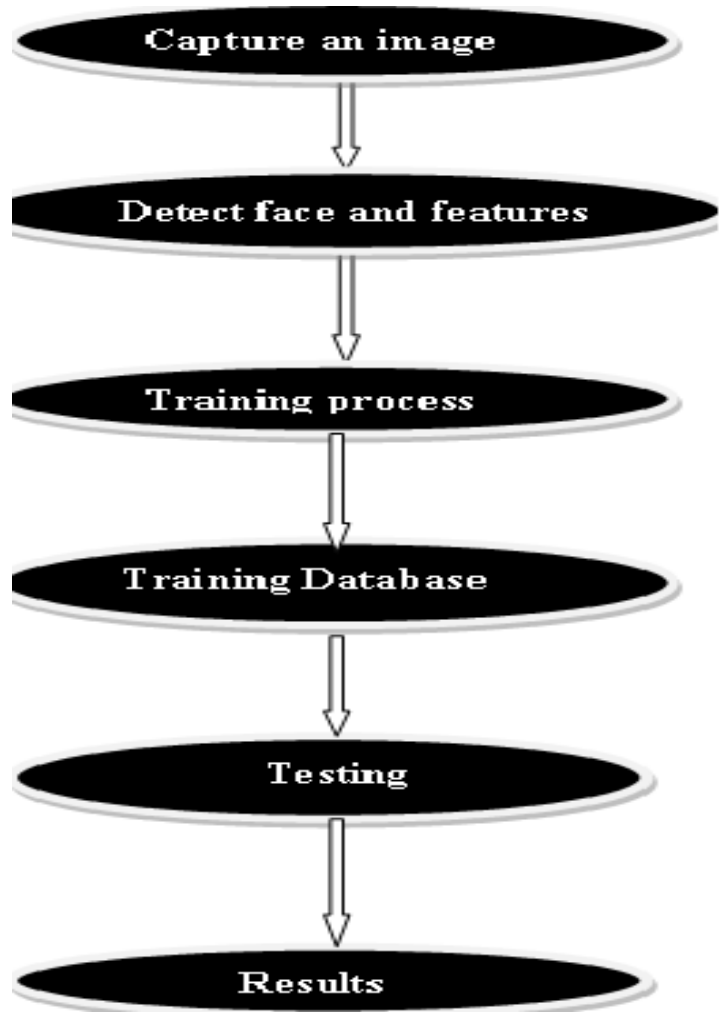

Figure 1: Flowchart for face recognition

## V. RESULTS AND DICUSSION

For carrying out the training procedure a three layer neural network is used. The training and testing of data has been done using the neural network tool in the MATLAB has been used. We have used feed forward networks under the supervised learning architecture of neural network tool box to compute our data , in which one way connections operate and no feedback mechanism is included.

Table 1: Sample Data

| S.NO | SAMPLE | INPUT DATA | TARGET DATA |
|---|---|---|---|
| 1 | 1 | 10010101 | 10001111 |
| 2 | 2 | 11001100 | 11110000 |
| 3 | 3 | 10101100 | 10101011 |

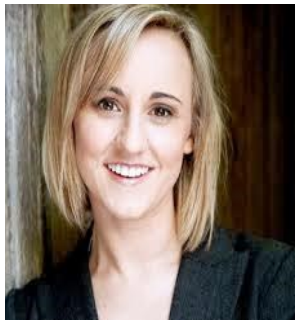

**Figure 2 Sample image 1**

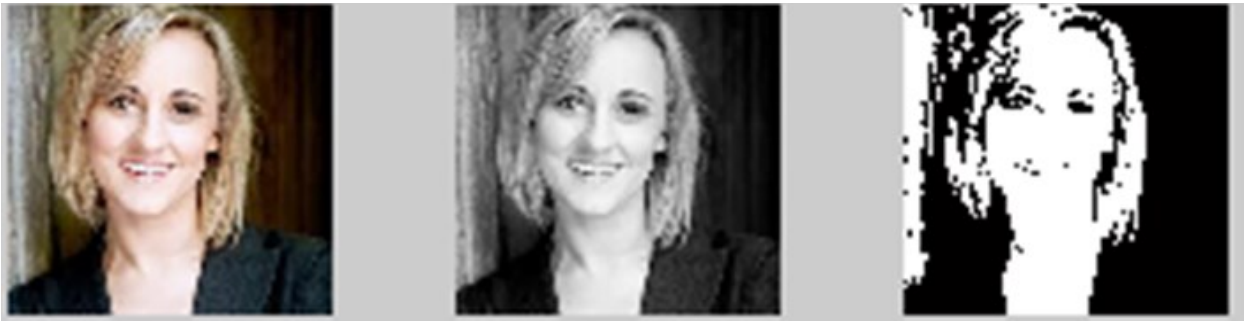

**Figure 3 sample image 1 in RGB, sample image 2 in grayscale and sample image 3 in binary form.**

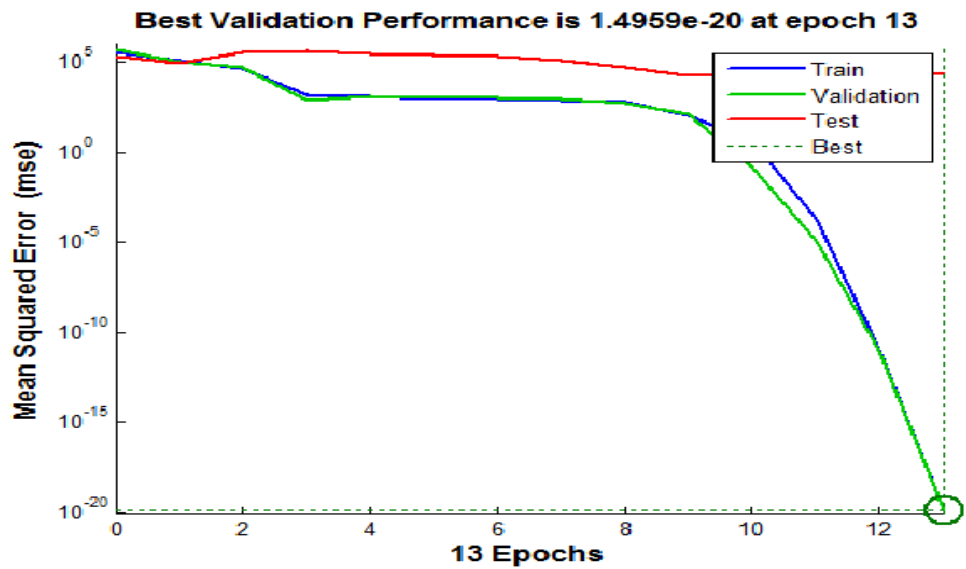

**Figure 4:Performance plot for sample data in table 1**

Table 1: Sample Data for sample image 2

| INPUT DATA | | | | TARGET DATA |
|---|---|---|---|---|
| 462 | 0 | 0 | 102 | 1100 |
| 342 | 0 | 0 | 78 | 0010 |
| 234 | 0 | 0 | 65 | 1001 |
| 500 | 0 | 0 | 132 | 1010 |
| 222 | 0 | 0 | 69 | 1011 |
| 165 | 0 | 0 | 45 | 0111 |

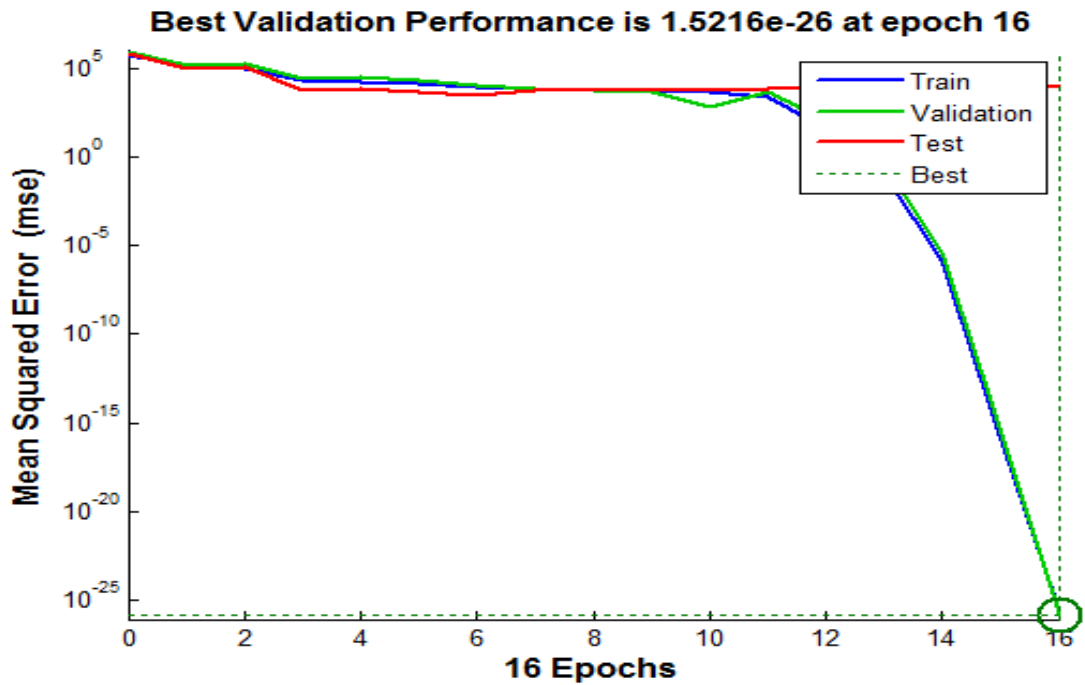

Figure 5: Performance Plot for sample data in table 2.

## VI. CONCLUSION

In the end we can conclude by stating that it is just another method employed for facial recognition where in a facial image was captured and the necessary steps to train and process the image were carried out. The training was performed 5-10 times to obtain appropriate results. This network has been used for processing various samples and proves to be accurate enough in providing results. The future scope can be done to automate the entire process or partially.